\documentclass[runningheads]{llncs}
\usepackage[utf8]{inputenc}

\usepackage[T1]{fontenc}
\usepackage{amsmath,array,color,multirow,enumitem,url,float}
\usepackage{amssymb}
\usepackage[ruled,vlined]{algorithm2e}
\usepackage{graphicx}

\begin{document}
\title{Machine Learning in Sports: A Case Study on Using Explainable Models for Predicting Outcomes of Volleyball Matches}
\titlerunning{Explainable Machine Learning in Sports}  
%
\author{Abhinav Lalwani\inst{1*} \and
Aman Saraiya\inst{2*} \and
Apoorv Singh\inst{2*} \and
Aditya Jain\inst{1*} \and
Tirtharaj Dash\inst{1*}}
\authorrunning{A. Lalwani et al.}
%
\institute{Department of CSIS, BITS Pilani, K.K. Birla Goa Campus, Goa, India \and
Department of EEE, BITS Pilani, K.K. Birla Goa Campus, Goa, India \\
\email{Abhinav Lalwani Email ID: f20180352@goa.bits-pilani.ac.in} \\
\email{Aman Saraiya E-mail ID: f20180454@goa.bits-pilani.ac.in}\\
\email{Apoorv Singh E-mail ID:
f20180640@goa.bits-pilani.ac.in} \\
\email{Aditya Jain E-mail ID:
f20180243@goa.bits-pilani.ac.in} \\
\email{Tirtharaj Dash E-mail ID: tirtharaj@goa.bits-pilani.ac.in}}
\maketitle              
\begin{abstract}
Machine Learning has become an integral part of engineering design and decision making in several domains, including sports. Deep Neural Networks (DNNs) have been the state-of-the-art methods for predicting outcomes of professional sports events. However, apart from getting highly accurate predictions on these sports events outcomes, it is necessary to answer questions such as "Why did the model predict that Team A would win Match X against Team B?" DNNs are inherently black-box in nature. Therefore, it is required to provide high-quality interpretable and understandable explanations for a model's prediction in sports. This paper explores a two-phased Explainable Artificial Intelligence(XAI) approach to predict outcomes of matches in the Brazilian volleyball League (SuperLiga). In the first phase, we directly use the interpretable rule-based ML models that provide a global understanding of  the  model's  behaviors  based  on  Boolean  Rule  Column  Generation  (BRCG;  extracts  simple  AND-OR  classification  rules)  and  Logistic  Regression  (LogReg;  allows  to  estimate  the  feature  importance scores). In the second phase, we construct non-linear models such as Support Vector Machine (SVM) and Deep Neural Network (DNN) to obtain predictive performance on the volleyball matches' outcomes. We construct the "post-hoc" explanations for each data instance using ProtoDash, a method that finds prototypes in the training dataset that are most similar to the test instance, and SHAP, a method that estimates the contribution of each feature on the model's prediction. We evaluate the SHAP explanations using the faithfulness metric. Our results demonstrate the effectiveness of the explanations for the model's predictions.

\keywords{Explainability (XAI) \and Machine Learning \and Sports Analytics \and Volleyball}
\end{abstract}
\section{Introduction}
\label{intro}
With the increase in the amount and the type of data collected for various sports and increased computational power, it is now possible to gain deeper insights into complex information about the matches. Machine Learning algorithms aim to assist coaches and sport managers, to track a player's or team's performance, possible player's injury, scouting, and making sport-betting decisions. The use of machine learning algorithms to predict the result of a future match can be of significant use to make the right betting decisions. All these factors drive the research on machine learning in sports. 

We collect data of the Brazil SuperLiga volleyball league, for all matches over the past 8 years, using web scraping. Appropriate processing and averaging methods are used to store the relevant features for every match. In the first place, we attempt to develop Machine Learning models for predicting the winner of the Volleyball matches based on the standings and the performances of the teams in the previous matches. For humans to trust the predictions made by these models it is required to have an understanding of the reason behind a particular prediction. So we explore two different approaches for obtaining explanations on the predictions made by the models. 
In the first phase, we investigate the use of directly interpretable models LogReg and BRCG. Post-hoc interpretation methods like SHAP and ProtoDash are investigated in the second phase. 

This paper is organized as follows. In section 2, relevant literature related to sports prediction and explainability are reviewed. Section 3 provides details of the prediction explanation methods used in this work. All the experiments carried out and the observations are reported in section 4. Finally, the conclusions drawn from the present study are detailed in section 5.

\section{Related Works}
One of the earliest studies to consider AI in the analysis of sports performance was done by Lapham and Bartlett in 1995~\cite{ArandaCorral2013ComplexCL}. They showed that methods from artificial intelligence could be a rewarding future direction for the discipline.
Since then, many different AI techniques have been applied for sports prediction.

Purucker~\cite{535226} attempted to predict results in the National Football League (NFL) using a Neural
Network Model. Kahn~\cite{articleKahn} built on the work of Purucker, achieving 75\% accuracy across the
matches of week 14 and 15 of the NFL. In 2015, Tax et al.~\cite{unknown} combined dimensionality
reduction techniques with Machine Learning algorithms to predict a Dutch football competition.
Herbinet et al. \cite{Herbinet} used a variety of machine learning techniques to predict the score and
outcome of football matches, using in-game match events. 

Although not as popular as soccer or football, volleyball is a team sport that has a national as
well as international level competitions in almost every country. 
Wenninger et al.~\cite{articleWenninger} predicted
results of beach volleyball using machine learning methods. T\"umer and Ko\c cer~\cite{doi:10.1080/24748668.2017.1331570} used artificial
neural networks to predict league rankings in volleyball. However, there has been little to no
prior research on predicting volleyball match results. 
\section{Prediction Explanation Methods}
Explanation about a prediction can be obtained by either using a directly interpretable prediction model or by using post-hoc interpretation methods on black-box models. This paper explores both approaches to get explainable predictions of the Volleyball matches. Boolean Decision Rule via Column Generation (BRCG) and Logistic Regression (LogReg) fall into the category of directly interpretable models. For post-hoc interpretation, we use ProtoDash and SHapley Additive exPlanations (SHAP) methods. BRCG and LogReg provide globally interpretable models, i.e. generalized classification approach for all the instances can be obtained. Whereas, ProtoDash and SHAP methods provide local interpretability, i.e. they provide prediction explanation only for the particular sample under consideration. This section describes the details of all the given explanation methods.

\subsection{BRCG}
The Boolean decision Rule via Column Generation (BRCG)~\cite{dash2020boolean} algorithm generates a directly interpretable model which has global interpretability. The algorithm produces a binary classifier in the disjunctive normal form (DNF; OR of ANDs) rule. This paper uses the BRCG-light algorithm, which uses the heuristic beam search~\cite{arya2019explanation}, to generate the classifier.





\subsection{LogReg}

Logistic regression models the probabilities for classification problems with a finite set of outcomes. Logistic Regression's goal is to learn a classifier that can make a decision about the class of a new observation. It solves this task by learning, from training data, a set of weights and a bias term. Each weight $w_i$ is associated with one of the input features $x_i$ . The absolute value of a weight $\lvert w_i \rvert$ represents the importance level of the feature $x_i$.

\subsection{SHAP}
SHapley Additive exPlanations (SHAP) \cite{lundberg2017unified} is a game-theoretic approach to get post-hoc interpretation on the prediction of any machine learning model. This model's goal is to explain the prediction for any instance xi as a sum of contributions from its
individual feature values. Shapley values are used to fairly distribute the contributions among the feature values. "Feature value" refers to the numerical or categorical value of a feature for the instance xi.

To compute the Shapley value of a particular feature $i$ it is required to look at all feature combinations where $i$ was included and where it was not included to obtain the prediction. With this, we can get the marginal contribution of $i$ ($MC_i$) for all feature combinations and then average them to get the total contribution of $i$ to the prediction value. This paper uses Kernel SHAP~\cite{lundberg2017unified}, to compute Shapley values.



We evaluate the SHAP explanations using the faithfulness metric \cite{NEURIPS2018_3e9f0fc9}.  This metric measures correlation between the importance assigned by SHAP to attributes and the effect of each of the attributes on the performance of the model.  The importance should be proportional to the effect on the performance.

\subsection{ProtoDash}
ProtoDash Algorithm \cite{gurumoorthy2019efficient} is a method for selecting prototypical examples that capture the distribution of dataset. It gives weight to each prototype to quantify data representation. The algorithm finds diverse prototypes  that reflects on the dataset differently to give complete picture of the dataset. Once the initial prototype is discovered, algorithm searches for the next prototype and whilst looking for an example with common behavior, it also try to discover new characteristics and ensures it is different from the first.

ProtoDash can be used to discover examples in the training dataset that best represent the distribution of a test input. Thus, this algorithm provides an intuitive method of understanding the underlying characteristics of any prediction made by the model.

\section{Empirical Evaluation}
\subsection{Aims}
Our aims in this section are:
\begin{enumerate}[noitemsep,topsep=0pt]
\item We intend to show that the black box models perform better than the white box models, thus, showing
that there is a trade off between interpretability and predictive accuracy.
\item We intend to show that the gap between accuracy and interpretability can be bridged by constructing
post hoc explanations for the black box models.
\item We intend to show that these explanations are sensible.
\end{enumerate}
In all experiments, the predictive performance of a model will mean accuracy on a test-set, and comparing
models means comparing these accuracy scores.

\subsection{Materials}
We scrape data from the website FlashScore.in to build our data set.  We collected data from the seasons 2010-11 to 2018-19 of the Brazilian Volleyball League (SuperLiga), totaling 1289 matches. While scraping the data, we store metadata for each team, so that for each match, we can use the metadata to import contextual data from the previous matches for each of the two teams playing the match. All the experiments are conducted in a Python environment in Google Colaboratory. We used the AIX360 library ~\cite{arya2019explanation} for building the BRCG, ProtoDash, and SHAP models.
\newline
For each match, we collect the following features for both teams, inspired by a study conducted by Tax et al. ~\cite{unknown} and the features used by McCabe et al. ~\cite{inproceeding} These features are human interpretable, which makes it easier to construct explanations for the predictions.

\begin{enumerate}
    \item \textbf{Matches won}: The win ratios of basketball teams were shown to be relevant features in predicting National Basketball League (NBA) basketball matches by Miljkovic et al.~\cite{inproceedings}.
    
    \item \textbf{Average points scored and conceded} : Baio et al. \cite{doi:10.1080/02664760802684177} proposed a Gamma distribution mixture model for Italian Serie A match prediction which used the number of goals scored and conceded by both teams. We use exponential averaging to estimate the average points scored and conceded, so as to give more importance to recent matches.
    
    \item \textbf{Performance in earlier matches} : Aranda-Corral et al. \cite{ArandaCorral2013ComplexCL} found the previous matches between the same teams to have medium to high correlation with the match result in their study focusing on the Spanish national soccer league. We encode this by calculating the exponential average of the set difference of the previous matches played between the two sides.
    
    \item \textbf{Form} : Goddard \cite{https://doi.org/10.1111/j.1740-9713.2006.00145.x} concluded that losing streaks result in an increased winning probability and winning streaks result in a decreased winning probability. We attempt to take into account whether the team is on a winning or losing streak by calculating the form in the previous 5 matches(this is estimated using exponential averaging of the set difference of previous matches).
    
    \item \textbf{Home/away form} : Palomino et al. \cite{article} showed that home advantage plays an important role in predicting match outcomes. Hence, we add separate columns for home and away form to our data, as this will estimate how good a team is playing at home/away from home.
    
    \item \textbf{Importance of Match} : Goddard \cite{https://doi.org/10.1111/j.1740-9713.2006.00145.x}, showed specific end-of-season matches to be significant for match outcome. We assign importance of 0 to league matches, 1 to quarter-final games, 2 to semi-final games and 3 to final games.
    
    \item \textbf{Rest} : Fatigue was used as one of the features by Constantinou et al. \cite{CONSTANTINOU2012322} in
their prediction model. We attempt to gauge fatigue as the number of days played since the previous match. A team having 7 days of rest is considered to be well-rested and thus if the number of days since the previous game is more than 7, we reduce it to 7.
    
    \item \textbf{Team Ranking} : The position of the team on the league ladder is based on a list of the teams. This feature's use is obvious as a high ranking team is expected to defeat a low ranking team.
    
    \item \textbf{Previous Year's Ranking} : We use this feature as a team who performed well in the previous year is expected to perform well in the next year as well.
    
    \item \textbf{Performance in Previous Game} :The performance of the team in their
most recent game. In the first round, this value is typically taken from the previous season's final game.
    
    \item \textbf{Result}(Output Parameter) : This is set as 1 if the home team won the match and it set as 0 if the away team won the match.

\end{enumerate}

\subsection{Method}
First, we split the dataset into train and test sets. Then, we build several different models using the training set. These models include directly-interpretable (white-box) models, BRCG and Logistic Regression,
which provide global understanding of their behaviour, and non-interpretable (black-box) models, Support Vector Machine, Neural Networks and Linear Discriminant Analysis. We then compare the predictive
accuracy of all the models on the test set. If the black-box models perform better than the white-box
models, we construct post-hoc, local explanations for the predictions made by these models, using the
SHAP and ProtoDash methods. For each match in the test set, ProtoDash returns a set of examples
that represent the match in test set, and SHAP returns Shapley values, which can be used to explain
the prediction as a sum of contributions of individual features.
\subsection{Results}
First, we compare the result of the white-box models (BRCG and LogReg) with black-box models (SVM, NN
and LDA). Results from experiments on all the models are tabulated in Table 1.

The rule learnt by BRCG is: Predict Y=1 if [Away team's last year position > 3.00 AND head-to-head form > -1.02 AND home team's last year position $\leq$   10.00 AND home team's win percentage >
10.53], else predict Y=0. Using logistic regression, we found that the most important features were: away team's current position, home team's average points and the away team's position in the table in the previous year.

The principal findings in Table 1 are these: (a) For all three metrics (Accuracy, F1-Score and AUCROC, black-box models are able to outperform white-box models. (b) SVM model provides the most
accurate predictions among all models. Thus, we use the SVM model for our further experiments on explainability.

\begin{table}[ht]
    \centering
    \caption{Results of prediction using  different models}
    \begin{tabular}{|l|l|l|l|l|}
    \hline
        Models & Type of Model & Accuracy & F1-Score & AUC-ROC \\ \hline
        SVM & Black-box & 0.7790 & 0.7971 & 0.7771 \\ \hline
        Artificial Neural Network & Black-box  & 0.7713 & 0.7870 & 0.7706 \\ \hline
        LinearDiscriminantAnalysis & Black-box & 0.7519 & 0.7681 & 0.7514 \\ \hline
        Logistic Regression & White-box & 0.7403 & 0.7581 & 0.7394 \\ \hline
    BRCG &  White-box & 0.7218 & 0.7446 & 0.7182 \\ \hline
    \end{tabular}
\label{tab:compression}
\end{table}

\begin{figure}[h]
     \centering
         \includegraphics[width=\textwidth]{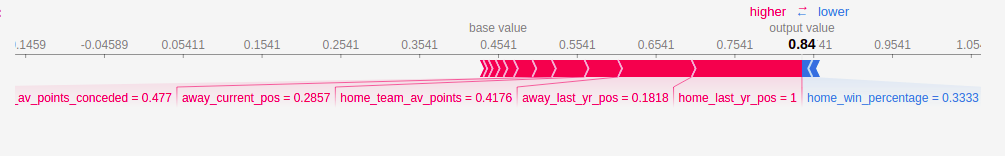}
        \caption{SHAP Explanations for the predictions made by SVM model for the match between Canoas and America Volei on 5/12/2014. The model predicts that Canoas will win the match with probability 0.84. The base value is 0.45, however the feature values of home team's position in previous season, away team's position in previous season, home team's average points, and others push the predicted value higher, and the value of home team's win percentage pushes the predicted value lower }
        \label{fig:SHAP}
\end{figure}

\noindent
\textbf{Can we provide post-hoc explanations for the black-box models?} We use two techniques, ProtoDash and SHAP, to provide local explanations. For any match in the test set, ProtoDash finds 5 similar examples from the training set, and assigns each match an importance weight. An example is shown in Table~\ref{tab:ProtoDash}. SHAP shows features contributing to push the model output from the base value (the average model output over the training dataset) to the actual model output. Features pushing the prediction lower are shown in blue, those pushing the prediction higher are in red. An example is shown in Fig~\ref{fig:SHAP}. The predictions made by SHAP achieve an average faithfulness score of 0.60 (on a scale of -1 to +1).

\noindent
\textbf{Do these explanations make sense?} We analyse the SHAP explanations using the example in Fig~\ref{fig:SHAP}. In Fig 1, we can see that the two features contributing most to the prediction that the home team wins are that the home team had a good position in the league in the previous season, and that the away team had a poor position in the previous season. This makes sense as a team who did well in the previous season is expected to do well in the next season as well. Also, the fact that the home team has a high value of average points scored, and the away team has a high value for average point conceded pushes the prediction towards the home team. This makes sense as the home team is expected to outscore the away team, and thus win the match. Average points scored and conceded have been shown to be good statistical indicators of match result by Baio et al.
\cite{doi:10.1080/02664760802684177}.

The ProtoDash predictions are analysed using the example in Table~\ref{tab:ProtoDash}.  Based on importance weight outputted by the method, we see that Prototype 1 is the most representative match by far. This is (intuitively) confirmed from the feature similarity table(Table~\ref{tab:ProtoDash}) where almost all of the features (17 out of 19) of this prototype are more than 50\% similar to  that  of  the  chosen  match  whose  prediction  we  want  to  explain.

\begin{table}[ht]
     \caption{Prototypes found by ProtoDash for the match between Canoas and America Volei on 5/12/2014. The numerical values in the table show the similarity of each feature in the prototypes with the feature value of the chosen match. The above table depicts the five closest matches in the training set to the chosen match in the test set.
    Looking at these examples which have had a similar result, gives any human more confidence in the prediction made by the model.}
    \resizebox{\textwidth}{!}{\begin{tabular}{|l|l|l|l|l|l|}
    \hline
        Feature & Prototype-1 & Prototype-2 & Prototype-3 & Prototype-4 & Prototype-5 \\ \hline
        Away team's current position & 0.77 & 0.26 & 0.59 & 0.2 & 0.45 \\ \hline
		Away team's position in previous season & 1 & 0.75 & 0.18 & 0.24 & 0.75 \\ \hline
		Away team's previous game performance & 1 & 0.08 & 1 & 1 & 1 \\ \hline
		Away team's average points & 0.61 & 0.23 & 0.39 & 0.22 & 0.95 \\ \hline
		Away team's average points conceded & 0.31 & 0.26 & 0.05 & 0.3 & 0.97 \\ \hline
		Away team's away form & 0.76 & 0.18 & 0.51 & 0.19 & 0.7 \\ \hline
		Away team's form & 0.76 & 0.1 & 0.72 & 0.11 & 0.89 \\ \hline
		Away team's win percentage & 0.96 & 0.14 & 0.81 & 0.25 & 0.52 \\ \hline
		Head to head form & 0.52 & 0.14 & 0.33 & 0.43 & 0.11 \\ \hline
		Home team's current position & 0.65 & 0.12 & 0.28 & 0.65 & 0.42 \\ \hline
		Home team's position in previous season & 0.75 & 0.08 & 1 & 0.57 & 0.14 \\ \hline
		Home team's previous game performance & 1 & 0.13 & 1 & 1 & 0.13 \\ \hline
		Home team's average points & 0.73 & 0.39 & 0.22 & 0.24 & 0.73 \\ \hline
		Home team's average points conceded & 0.57 & 0.88 & 0.16 & 0.66 & 0.34 \\ \hline
		Home team's form & 0.62 & 0.48 & 0.91 & 0.14 & 0.33 \\ \hline
		Home team's home form & 0.6 & 0.16 & 0.45 & 0.29 & 0.7 \\ \hline
		Home team's rest time & 0.19 & 0.19 & 0.19 & 0.19 & 0.43 \\ \hline
		Home team's win percentage & 0.67 & 0.21 & 0.69 & 0.41 & 0.21 \\ \hline
		Match importance & 1 & 1 & 1 & 0.08 & 1 \\ \hline
		Weight & 0.694565 & 2.95E-05 & 0.164856 & 0.089472 & 0.0510778 \\ \hline
		Home Team & Corinthians & Minas & America Volei & Cimed & Sao Bernardo \\ \hline
		Away Team & Maringa & Sada Cruzeiro & Funvic & Volei Futuro & Sesi \\ \hline
		Date & 20-02-2019 & 15-01-2014 & 11-01-2014 & 01-03-2011 & 01-01-2014 \\ \hline
    \end{tabular}}
\label{tab:ProtoDash}
\end{table}

\section{Conclusion}
In this paper, we have examined the use of both black-box and white-box models for predicting results
of volleyball matches. Our findings suggest that black-box models are able to achieve a better predictive
accuracy than white-box models. We have also shown that we can construct post-hoc local explanations
of the predictions made by the black-box models, using methods such as SHAP and ProtoDash. Further,
we have examined that the explanations produced by these models are sensible. 

The feature-set we have used is very general, however a limited amount of data is available for each
league. Hence, meta-learning methods may be able to achieve better performance, as they can learn
across several tasks. These tasks could be from different leagues of the same sport, or even different
sports. We leave this as future work

\section{Acknowledgements}
We would like to thank Teaching Assistants of the course BITS F464 (Machine Learning): Het Shah, Avishreee Khare and Aditya Ahuja, as well as Prof. Ashwin Srinivasan for their valuable guidance
during this project.

%
%
%
\bibliographystyle{splncs04}
%

\bibliography{main}

\end{document}